\documentclass[tinyml]{acmart}

\usepackage{fancyvrb}
\AtBeginDocument{%
  \providecommand\BibTeX{{%
    \normalfont B\kern-0.5em{\scshape i\kern-0.25em b}\kern-0.8em\TeX}}}

\setcopyright{rightsretained}
\copyrightyear{2022}
\acmYear{2022}

\begin{document}

\title{How to Manage Tiny Machine Learning at Scale – An Industrial Perspective}

\author{Haoyu Ren}
\email{haoyu.ren@siemens.com}
\orcid{0000-0002-0241-6507}
\affiliation{%
  \institution{Siemens AG}
  \institution{Technical University of Munich}
  \city{Munich}
  \state{Bavaria}
  \country{Germany}
}

\author{Darko Anicic}
\orcid{0000-0002-0583-4376}
\email{darko.anicic@siemens.com}
\affiliation{%
  \institution{Siemens AG}
  \city{Munich}
  \state{Bavaria}
  \country{Germany}
}

\author{Thomas A. Runkler}
\orcid{0000-0002-5465-198X}
\email{thomas.runkler@siemens.com}
\affiliation{%
  \institution{Siemens AG}
  \institution{Technical University of Munich}
  \city{Munich}
  \state{Bavaria}
  \country{Germany}
}

\begin{abstract}
Tiny machine learning (TinyML) has gained widespread popularity where machine learning (ML) is democratized on ubiquitous microcontrollers, processing sensor data everywhere in real-time. To manage TinyML in the industry, where mass deployment happens, we consider the hardware and software constraints, ranging from available onboard sensors and memory size to ML-model architectures and runtime platforms. However, Internet of Things (IoT) devices are typically tailored to specific tasks and are subject to heterogeneity and limited resources. Moreover, TinyML models have been developed with different structures and are often distributed without a clear understanding of their working principles, leading to a fragmented ecosystem. Considering these challenges, we propose a framework using Semantic Web technologies to enable the joint management of TinyML models and IoT devices at scale, from modeling information to discovering possible combinations and benchmarking, and eventually facilitate TinyML component exchange and reuse. We present an ontology (semantic schema) for neural network models aligned with the World Wide Web Consortium (W3C) Thing Description, which semantically describes IoT devices. Furthermore, a Knowledge Graph of 23 publicly available ML models and six IoT devices were used to demonstrate our concept in three case studies, and we shared the code and examples to enhance reproducibility: \href{https://github.com/Haoyu-R/How-to-Manage-TinyML-at-Scale}{Github}
  
\end{abstract}


\maketitle

\section{Introduction}
Recently, Tiny Machine Learning (TinyML), which concerns the execution of machine learning (ML) models on low-power microcontrollers found in Internet of Things (IoT) devices, has evolved rapidly. Shifting and distributing ML to IoT devices provide ML functionalities close to data sources without consistently depending on the internet connection. TinyML preserves data privacy and improves latency and energy efficiency. According to~\cite{statista2021}, over 25 billion microcontrollers (MCUs) were shipped in 2019 with a continuously growing demand. By bringing ML on all-pervasive low-power embedded devices, we can dramatically boost the intelligence of IoT applications. There have been significant advancements in developing TinyML algorithms and miniaturizing low-power IoT devices. 

Nevertheless, the heterogeneity of the TinyML ecosystem severely limits effective management and deployment of TinyML at scale, especially in industry settings, where an industrial IoT network involves hundreds or thousands of devices. Embedded devices are typically customized to a specific use case with limited capabilities to optimize computation and energy efficiency. They bear a variety of sensors, RAM and Flash sizes, computational power, and even runtimes.  Additionally, developing a TinyML model requires dozens of decisions, such as the neural network type, combination of layers, and neurons size, to improve its performance toward various working conditions. Also, preprocessing and postprocessing of each neural network model are individually crafted since an ML model only provides a meaningful output for correctly formulated input sensor data. Moreover, TinyML developers usually distribute ML models without clear/standard documentation of their functionalities and usage. These challenges lead to a highly fragmented ecosystem.

The proliferation and diversification of TinyML components (ML models and embedded devices) make managing a TinyML system in the industry a challenge. Questions to be addressed include: How to find which IoT devices in the field have the necessary sensors and capability to support a given ML model without examining each device individually? How to discover existing neural network models to be executed on a given embedded device without spending time reinventing a new model? How to manage and reuse TinyML components at scale? How to orchestrate the discovery result into an IoT application? 

To address these challenges, we present a novel concept for scaled management of TinyML by leveraging Semantic Web technologies. Here, we introduce an ontology (semantic schema) for describing ML models in the context of TinyML, including neural network representation, hardware requirements, and metadata. Furthermore, we propose the application of the World Wide Web Consortium (W3C) Web of Things (WoT)~\cite{W3C2021} to enrich IoT devices using standardized Thing Description (TD) ontology~\cite{Charpenay2016}, which enables the semantic description of IoT devices, such as the hardware specifications. We design the TinyML model ontology aligned with TD ontology to enable matching model requirements with IoT device capabilities. Thus, heterogeneous knowledge about TinyML components can be rendered in a unified language against the corresponding ontology and centrally hosted in a Knowledge Graph (KG), which provides advantages, such as vendor-agnostic knowledge management and reasoning over data. Also, SPARQL\footnote{\url{https://www.w3.org/TR/rdf-sparql-query/}} can be used to flexibly discover and interoperate TinyML components, which are made searchable and accessible like Web resources to everyone, thereby facilitating TinyML system management and strengthening cross-sector collaboration.

We aim to provide means for the community to build, manage, and leverage a TinyML KG. Code and examples are available for automatically generating the semantic representation given any neural network model with the format ".tflite" supported by TensorFlow Lite for Microcontrollers. We have not discussed the semantic representation of IoT devices in detail since TD ontology is standardized and well-established. Nevertheless, we have provided six semantic descriptions of IoT devices in our repository. Furthermore, we illustrated how to orchestrate TinyML components and deploy ML models to embedded devices upon knowledge matching. We do not expect every TinyML engineer to know semantics or to want to invest time in writing SPARQL queries. Therefore, we demonstrate how the management of TinyML in the industry could look like in the future by leveraging low-code platforms.

The remainder of the paper is organized as follows: Section \ref{section:sota} reviews related work on TinyML and IoT management. Section \ref{section:approach} describes our concept, starting from the TinyML neural network ontology design to building a KG that stores the TinyML system information, and ends with evaluating the concept with three case studies. Section \ref{section:framework} summarizes the workflow and generalizes the TinyML semantic management system. We showed the integration of the framework into a low-code platform. Finally, Section \ref{section:conclusion} presents the conclusion of the paper and discusses future work.

\section{Related Work}
\label{section:sota}

\subsubsection*{Proliferation of TinyML}

TinyML is surging across research and industrial communities. Recently, a survey~\cite{Shafique2021} summarized breakthroughs and challenges for promoting embedded ML. Better algorithms~\cite{Ren2021}~\cite{Cai2020} are designed to more efficiently use resources, improve model performance, and optimize the deployment in a real-world environment. Also, ultra-low-power AI chips~\cite{Giordano2021} and accelerators~\cite{Jiao2021} have been proposed to support always-on ML capability for an extended period by a battery. However, a joint design of hardware and algorithm~\cite{Sudharsan2021}~\cite{Jain2021}  is required to squeeze the performance since TinyML delivers ML solutions to constrained devices with limited resources.  Moving a step forward, frameworks for characterizing and assessing ML deployment on the edge~\cite{Qiu2021}~\cite{Corneliou2021} help us systematically tackle potential issues. Many platforms and resources, such as open online courses~\cite{Reddi2021}, X-CUBE-AI~\cite{STM21} from STM, Apache's TVM~\cite{Tianqi2018}, and TensorFlow Lite for Microcontrollers~\cite{David2020} from Google, are made available to accelerate TinyML. Besides, remarkable applications of TinyML show up across all fields, see~\cite{Ren2021a}~\cite{Herath2021}~\cite{Zhou2021}.

\subsubsection*{Management of TinyML} Increasing transparency and interoperability of TinyML components address heterogeneity in the ecosystem and ease management. Open Neural Network Exchange (ONNX)~\cite{Facebook2021} is an open-source project dedicated to sharing ML models across different frameworks. Also, TensorFlow Lite Metadata~\cite{tflitemeta2021} provides a schema for documenting ML models, for example, input/output information.  Another standard from Google for describing ML models that focus on expressing model constraints and performance is Model Card~\cite{Mitchell2019}. However, these solutions offer standalone documentation manually drafted accompanying each ML model and are hardly exchangeable and interoperable. Besides, we are less interested in converting ML model format, but rather their reusability incorporating IoT devices regarding TinyML.

Lately, the management of ML models has received increased attention. Authors in~\cite{Schelter2018} introduced challenges of model management in a set of ML use cases. Similarly, authors in~\cite{Vartak2016} described a database for tracking ML models. A KG called FAIRnets~\cite{Nguyen2020} collects information about publicly available neural networks and makes them exchangeable. Although all methods mentioned above provide certain degrees of expressiveness about ML models, they fall short in TinyML. To deal with the overwhelming complexity in TinyML, both ML models and hardware specifications must be investigated. 

Semantic Web technology is a good candidate for modeling heterogeneous knowledge about TinyML components that are human- and machine-understandable and has been effectively applied in IoT domain~\cite{Pandey2021}. The WoT technique~\cite{W3C2021} is introduced to make heterogeneous IoT devices accessible as agnostic as Web resources. Furthermore, a formal information schema is proposed by an W3C WoT Interest Group to semantically describe the capabilities and interfaces of things, known as TD ontology~\cite{Charpenay2016}. 

However, there remains a gap still between IoT devices and ML models. Challenges of managing TinyML components at scale, such as discovering appropriate combinations and initiating (mass) deployment, remain unsolved. A semantics-based approach \cite{Ren202201} for managing general on-device applications on distributed IoT devices is proposed. We specialized this approach and developed a neural network ontology for TinyML together with TD. Established semantic vocabularies are reused to maximize interoperability: The Semantic Sensor Network ontology (SSN)~\cite{Compton2012} describes sensors and their interactions incorporating the Sensor, Observation, Sample, and Actuator (SOSA) ontology~\cite{Janowicz2019}. The Semantic Smart Sensor Network (S3N) ontology~\cite{Rebai2021} was built on SSN to interpret smart sensors and their properties. Another semantic model expressing connected things in IoT is  \textbf{iot.schema.org}, an extension to \textbf{schema.org} that enriches structured data across the internet.

Finally, TinyML as a service (TinyMLaaS)~\cite{Doyu2021} supports the deployment of ML models on versatile embedded devices by building an abstract layer on top of various hardware compilers and automatically generating a compiled ML model for any given target hardware platform. It can serve as a supplement to the deployment step in our workflow.

\section{Approach}
\label{section:approach}

This section identifies the necessary information for building an ontology of neural network models regarding TinyML and elaborates on the semantic schema in detail. Since TD was used to express the capabilities of IoT devices, the TinyML model ontology was designed with the compatibility of TD in mind. Next, we introduced a KG containing 23 publicly available TinyML models and six IoT devices to demonstrate the semantic management framework. Finally, we presented three case studies to show the advantages of our approach.

\subsection{Building a TinyML Model Ontology}

\begin{figure*}[t]
      \includegraphics[width=0.95\linewidth]{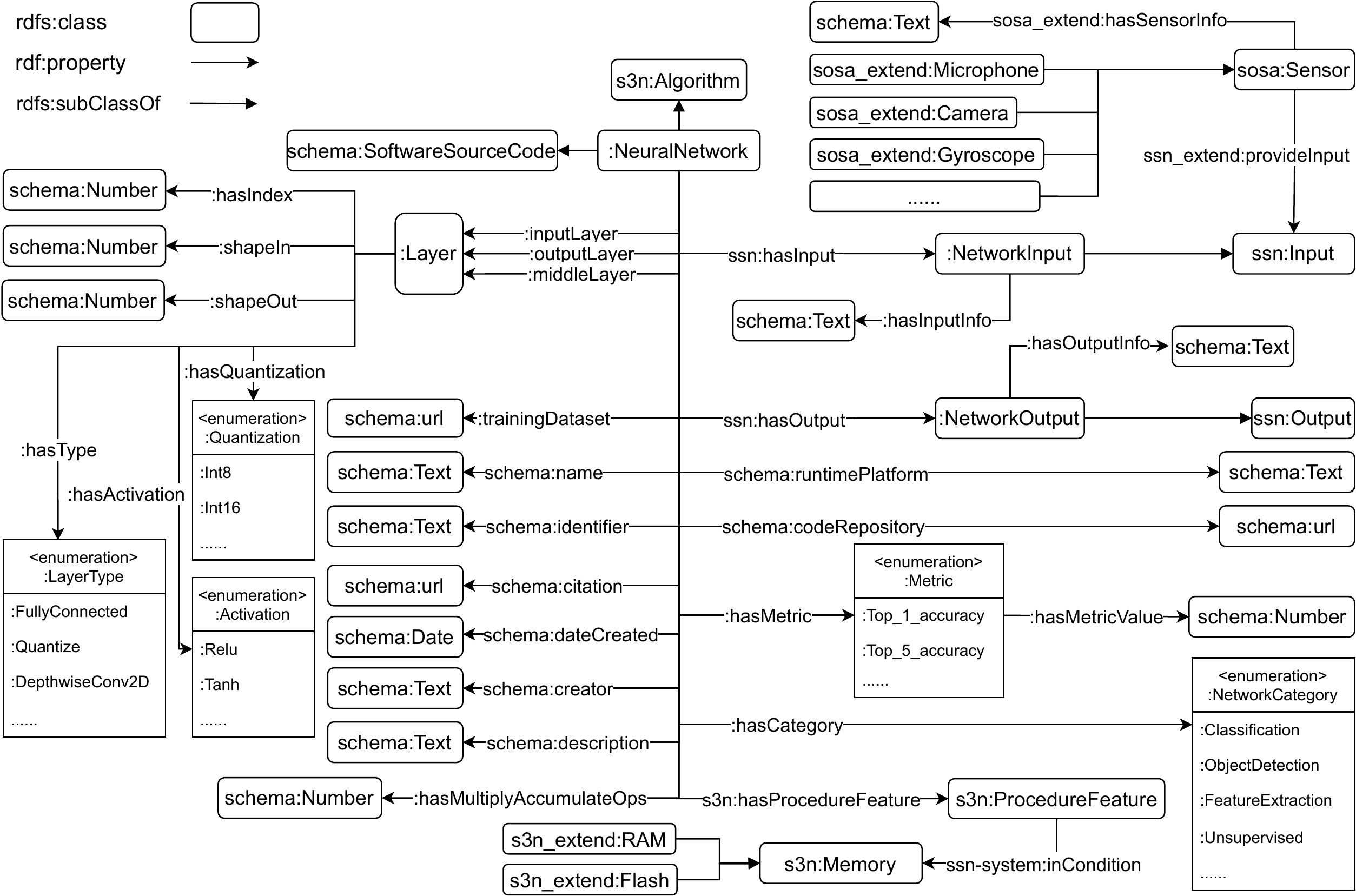}
      \caption{Tiny machine learning neural network ontology.}
      \label{neuralNetwork}
\end{figure*}

From Ontology 101~\cite{Noy2001}, we determined the granularity and scope of the ontology by listing five Competency Questions:

\begin{enumerate}
    \item How can we describe an ML model running on an embedded device?
    \item What metadata of a model should be included? For example, creator, identifier, and reference.
    \item How can we manifest the standard features of an embedded ML model? For instance, a combination of different layers, quantization, and input/output information.
    \item What hardware-related information should be considered? For example, memory and sensor requirements.
    \item How should we design the vocabulary that is aligned with TD?
\end{enumerate}

Authors in~\cite{Noy2001} suggested that creating a new ontology by reusing existing ones improves interoperability and avoids reinventing the wheel. Thus, we used S3N ontology~\cite{Rebai2021} as the upper layer, which is built on top of the SOSA~\cite{Janowicz2019}, SSN~\cite{Compton2012}, and W3C WoT TD~\cite{Charpenay2016} ontologies. The choice of S3N contributes compatibility with other Web standards and enhances the semantic alignment between the hardware and ML models, namely, TD and the proposed neural network ontology — this answers the Competence Question (5).

The prefixes and the namespaces of the used ontologies are given as follows:

\begin{Verbatim}[fontsize=\small] 
    # RDF Vocabulary
    rdf: <http://www.w3.org/1999/02/22-rdf-syntax-ns#> .
    # RDF Schema Vocabulary
    rdfs:<http://www.w3.org/2000/01/rdf-schema#> .
    # Schema.org Vocabulary
    schema: <https://schema.org> .
    # Units of Measure Vocabulary
    om: <http://www.ontology-of-units-of
         -measure.org/resource/om-2/> .
    # SSN Ontology
    ssn: <http://www.w3.org/ns/ssn/> .
    # S3N Ontology
    s3n: <http://w3id.org/s3n/> .
    # SOSA Ontology
    sosa: <http://www.w3.org/ns/sosa/> .
    # Thing Description Ontology
    td:  <http://www.w3.org/ns/td#> .
    # Extension of the SOSA ontology
    sosa_extend:  <http://tinyml-schema.org/sosa_extend#> .
    # Extension of the SSN ontology
    ssn_extend:  <http://tinyml-schema.org/ssn_extend#> .
    # Extension of the S3N ontology
    s3n_extend:  <http://tinyml-schema.org/s3n_extend#> .
\end{Verbatim}

\begin{figure*}[t]
      \includegraphics[width=0.95\linewidth]{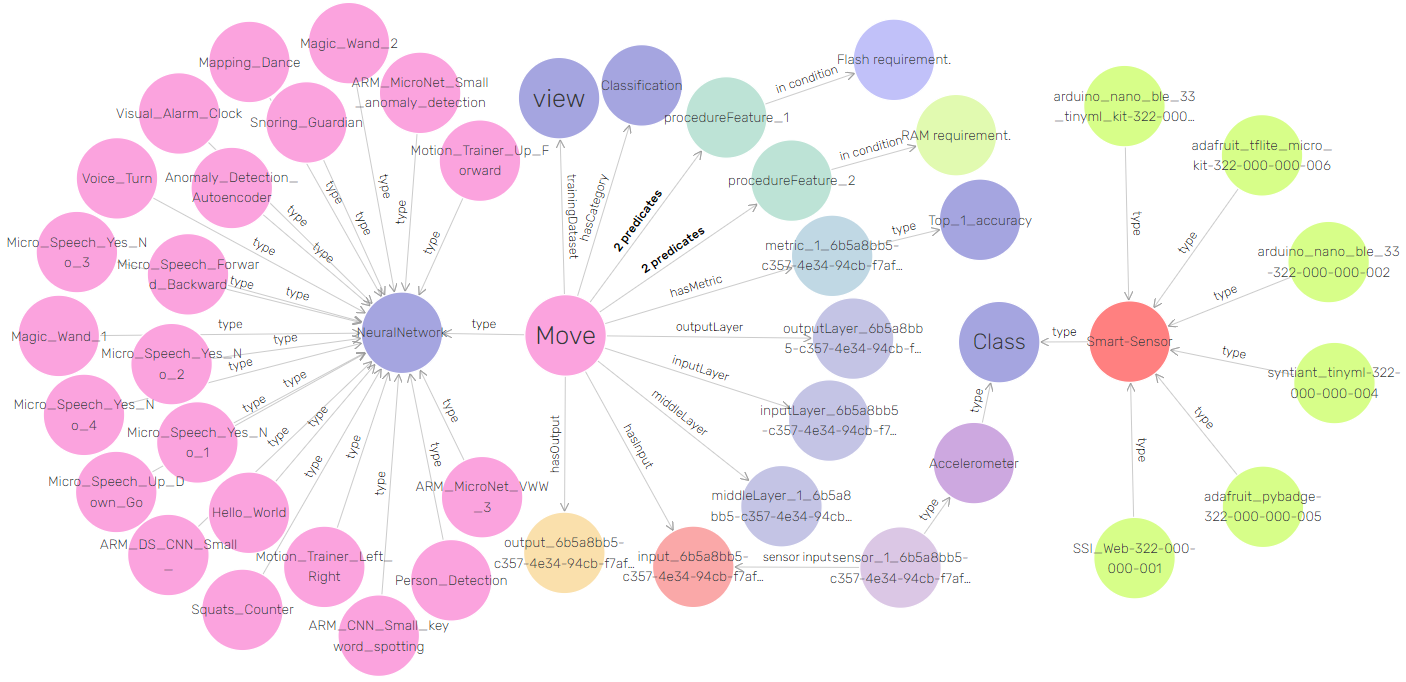}
      \caption{Tiny machine learning Knowledge Graph composed of 23 ML models and six IoT devices (simplified).}
      \label{kg}
\end{figure*}

Figure \ref{neuralNetwork} shows the semantic schema\footnote{We use the prefix \textbf{nnet:} to identify the namespace of our neural network ontology: <http://tinyml-schema.org/networkschema\#>} for TinyML models. Here, neural network is defined as a subclass of "s3n:Algorithm" and "schema:SoftwareSourceCode." Also, metadata can be added to enrich an ML model. For instance, \emph{identifier} is an ID used to identify a model uniquely. The property \emph{citation} allows a URL to be associated with a reference site, such as a paper or repository. Similarly, a ML model can have a \emph{code repository} indicating where the model can be retrieved, a \emph{training dataset} that contains information on the training dataset, the \emph{created date}, the \emph{creator} of the model, a human-readable \emph{description}, and so forth. These attributes make it possible to search ML models by various criteria and answer the Competency Question (2).

Furthermore, a neural network consists of various \emph{layers} broken into three categories: \emph{input}, \emph{output}, and \emph{middle} (hidden) layers. Each layer contains individual information, e.g., \emph{input shape}, \emph{output shape}, and \emph{quantization}. Additionally, a layer is defined by \emph{type}, such as \emph{FullyConnected}, \emph{Quantize}, \emph{DepthwiseConv2D}, thereby allowing type-specific information to be complemented. The structure of a ML model is represented by \emph{indexing} the layers. These answer the Competency Question (3).

Also, model-specific information is covered in the ontology, which is essential for gaining insight into an ML model and relating the model to hardware. A model can have an accuracy \emph{metric} that indicates its performance on a given \emph{training dataset}. However, accuracy is not the only factor to consider when selecting a model concerning TinyML. The counts of \emph{multiply-accumulate operations} can indicate a model's latency, which varies linearly~\cite{Banbury2020}. The \emph{RAM} and \emph{Flash} requirements are critical to assess whether a model fits into an IoT device and are described as subclasses of \emph{s3n:Memory} to conform to the TD. The \emph{runtime platform} provides the format and supporting framework information of a model. Additionally, different \emph{sensors} and relevant information are assigned to a model for imposing sensor requirements. Sensors can be divided into different types: a \emph{camera}, \emph{microphone}, \emph{accelerometer}, etc., which defines the input source. The \emph{category} and \emph{input/output} properties supply knowledge about the model use. Besides, the \emph{input} and \emph{output} of an ML model are denoted as subclasses of \emph{ssn:Inout} and \emph{ssn:Output} — these answer the Competency Questions (1) and (4). 

\subsection{Constructing a TinyML Knowledge Graph}

The ML model ontology was carefully created considering the integration with TD, which we applied to describe embedded devices. Due to Semantic Web technologies, information about both ML models and IoT devices can be semantically modeled against the respective ontology in a structured way and stored in the same KG for use in various IoT applications. We will not dive deep into the modeling of IoT devices with TD ontology, as it is well documented\footnote{\url{https://www.w3.org/WoT/}}. However, interested readers can find examples of semantic descriptions of six IoT devices in our repository as reference.

To validate the ontology and demonstrate the benefits of our approach, we presented a KG consisting of 23 publicly available neural network models and six different IoT boards, providing good coverage of variance in the TinyML community. Our target IoT devices are typically equipped with different sensors, hundreds of kilobytes of RAM, and a few megabytes of Flash. We collected TinyML models smaller than 1 MB trained using the Tensorflow Lite Microcontroller library. This embedded library is open-source, widely used in the community, well maintained, and frequently updated, improving the acceptance and propagation of our concept. Nevertheless, our solution is scalable to more powerful IoT devices, bigger models, and other frameworks.   

Scripts and examples of information extraction and knowledge formulation based on the proposed ontology are available in our repository. The semantic representation of a given ML model can be automatically generated along with a few user inputs since not all necessary information is acquired by parsing the model, such as citations, required sensors, etc.

Also, the KG can be visualized intuitively using a suitable tool. We use GraphDB\footnote{\url{https://www.ontotext.com/products/graphdb/}} to present it, as depicted in Figure \ref{kg}. A total of 23 individual ML models are shown on the left side of the figure. One model called "Move" is expanded in the center of the figure. Besides, six embedded devices are shown on the right side of the figure. Due to space limitations, other properties and connections are omitted.

Graph databases offer advantages in presenting relationships within a complex TinyML system. The flexibility of the KG allows easy component management and extends the existing semantic schema if needed. 

\subsection{Case Study}

Three SPARQL queries were used to show how other TinyML developers benefit from our concept when addressing various management problems. Note that the KG introduced above was used only for demonstration. Depending on the network scale, a KG in industrial IoT should contain more entries.

Suppose we trained a TinyML model for a motion classification task using accelerometer and gyroscope data. From the parsing result of the model, its minimum RAM and Flash requirements are 116 and 531 kb, respectively. There are many boards in the field that are heterogeneous in terms of built-in sensors and available memory. How can we find which boards are able to load this model without examining the IoT devices sequentially? As shown below, query 1, along with a corresponding formatted result, answers the question: 

\begin{Verbatim}[fontsize=\small, baselinestretch=0.75]
    Query 1:
    SELECT ?Board ?RAM ?Flash
    WHERE {
        ?Board a s3n:SmartSensor ;
            ssn:hasSubSystem ?system_1 ;
            ssn:hasSubSystem ?system_2 ;
            ssn:hasSubSystem ?system_3 .
        ?system_1 a sosa_extend:Accelerometer .
        ?system_2 a sosa_extend:Gyroscope .
        ?system_3 a s3n:MicroController ;
            s3n:hasSystemCapability ?x .
        ?x ssn-system:hasSystemProperty ?cond_1 .
        ?x ssn-system:hasSystemProperty ?cond_2 .
        ?cond_1 a s3n_extend:RAM ;
            schema:value ?RAM ;
            schema:unitCode om:kilobyte .
        ?cond_2 a s3n_extend:Flash ;
            schema:value ?Flash ;
            schema:unitCode om:kilobyte .
        FILTER (?RAM >= 116)
        FILTER (?Flash >= 531)
    }
    
    Result: 
    Board: 002; RAM: 172 kb; Flash: 628 kb.
    Board: 003; RAM: 187 kb; Flash: 785 kb.
\end{Verbatim}

Note that other constraints can be added for a more precise filtering result — for example, a specific runtime platform. 

Next, we want to empower a customized IoT board with ML functionality. Certain information is known: The board has a camera sensor, the available RAM and Flash memories are 127 and 576 kb, respectively, according to the feedback from the operating system. How do we find suitable ML models in the model zoo that runs on this board instead of spending much time creating a new model from scratch? We can use the following query:

\begin{Verbatim}[fontsize=\small, baselinestretch=0.75]
    Query 2:
    SELECT ?uuid ?MACs ?RAM ?Flash
    WHERE {
        ?nn a nnet:NeuralNetwork ;
            schema:identifier ?uuid ;
            ssn:hasInput ?input;
            nnet:hasMultiplyAccumulateOps ?MACs ;
            s3n:hasProcedureFeature ?x_1 ;
            s3n:hasProcedureFeature ?x_2 .
        ?x_1 ssn-system:inCondition ?cond_1 .
        ?x_2 ssn-system:inCondition ?cond_2 .
        ?cond_1 a s3n_extend:RAM ;
            schema:minValue ?RAM ;
            schema:unitCode om:kilobyte .
        ?cond_2 a s3n_extend:Flash ;
            schema:minValue ?Flash ;
            schema:unitCode om:kilobyte .
        ?sensor ssn_extend:provideInput ?input;
            a sosa_extend:Camera .
        FILTER (?RAM <= 127)
        FILTER (?Flash <= 576)
    }
    
    Result: 
    uuid: e3... ; MACs: 7387976; RAM: 116 kb; Flash: 531 kb.
    uuid: 10... ; MACs: 7158144; RAM: 94 kb;  Flash: 236 kb.
\end{Verbatim}

Finally, assume that we want to browser available ML models for a specific task, namely, speech recognition for "yes/no" commands. Meanwhile, we benchmark the results in terms of accuracy on a particular dataset, latency, and memory requirements. The query is structured as follows:

\begin{Verbatim}[fontsize=\small, baselinestretch=0.75]
    Query 3:
    SELECT ?uuid ?Category ?Acc ?MACs ?Min_RAM ?Min_Flash
    WHERE {
        ?nn a nnet:NeuralNetwork ;
            schema:identifier ?uuid ;
            schema:description ?description;
            ssn:hasInput ?input;
            nnet:hasCategory ?Category ;
            nnet:hasMetric ?metric ;
            nnet:hasMultiplyAccumulateOps ?MACs ;
            nnet:trainingDataset ?dataset ;
            s3n:hasProcedureFeature ?x_1 ;
            s3n:hasProcedureFeature ?x_2 .
        ?metric a nnet:Top_1_accuracy .
        ?metric nnet:hasMetricValue ?Acc .
        ?x_1 ssn-system:inCondition ?cond_1 .
        ?x_2 ssn-system:inCondition ?cond_2 .
        ?cond_1 a s3n_extend:RAM ;
            schema:minValue ?Min_RAM ;
            schema:unitCode om:kilobyte .
        ?cond_2 a s3n_extend:Flash ;
            schema:minValue ?Min_Flash ;
            schema:unitCode om:kilobyte .
        ?sensor ssn_extend:provideInput ?input;
            a sosa_extend:Microphone .
        FILTER regex(?description, "yes/no", "i")
        FILTER regex(str(?dataset), "speech_commands", "i")
    }ORDER BY ?Acc
        
    Result: 
    uuid: 3e...; Category: nnet:Classification; Acc: 0.82; 
    MACs: 381824; Min_RAM: 8.8 kb; Min_Flash: 8.9 kb.
    ... (Other entries are omitted to conserve space)
\end{Verbatim}

Users can discover various information and filter the results as efficiently as browsing the Web by constructing different queries. This enables an innovative way to manage a fragmented TinyML system and reduce the overhead of developing TinyML applications. 

\section{Semantic Management Framework}
\label{section:framework}

This section summarizes the workflow, generalizes the proposed semantic management system for TinyML, and shows how the concept can be integrated into a low-code platform, accelerating the development of IoT applications across the community.

\begin{figure}[t]
  \centering
  \includegraphics[width=\linewidth]{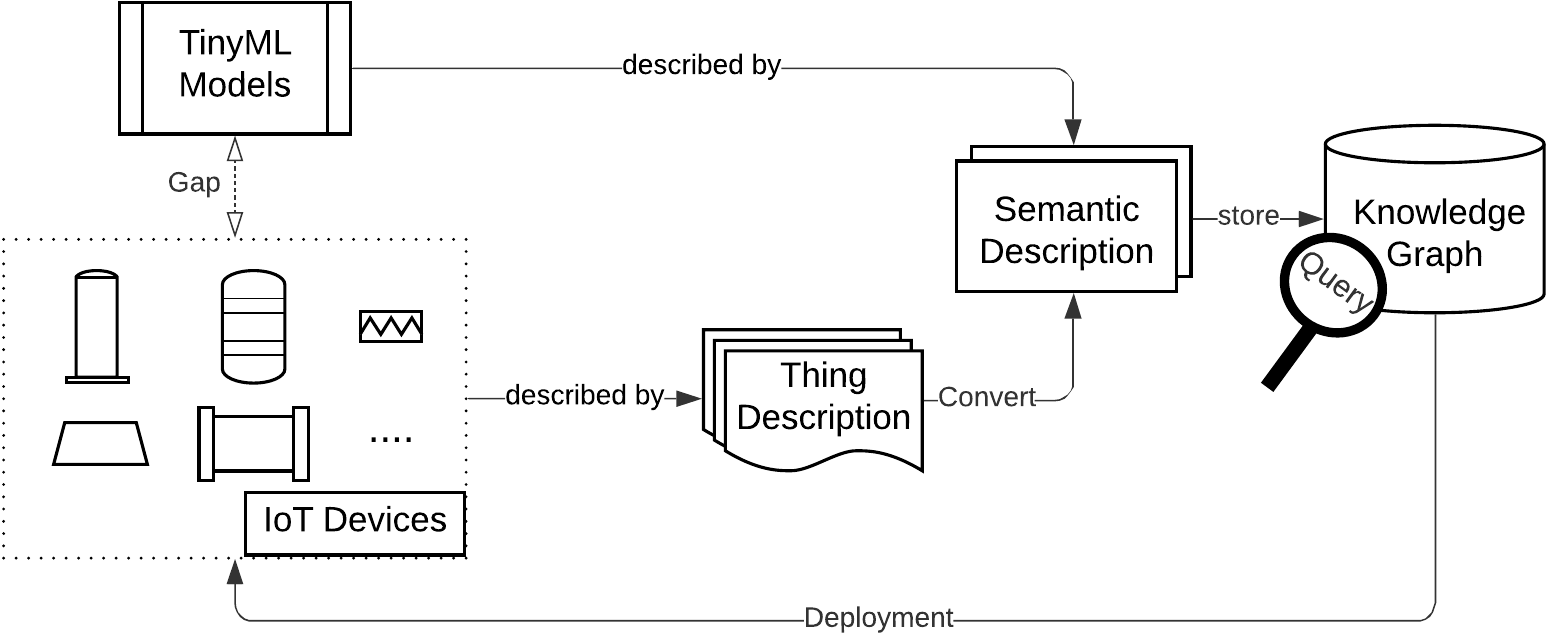}
  \caption{Framework for semantic management of tiny machine learning.}
  \label{framework}
\end{figure}

Figure \ref{framework} presents the system design with the framework comprising three building blocks: the semantic schemas (the TinyML model and TD ontologies), a KG, and deployment toolchain. The left of the figure illustrates the problem in the TinyML world: a barrier between heterogeneous embedded devices and various available ML models, resulting in a highly fragmented ecosystem challenging to manage. Thus, Semantic Web technologies were applied to tackle the problem. Information on ML models and IoT devices can be translated into semantic descriptions using the proposed ontologies. The semantic knowledge can be stored in a graph database, allowing us to query various information across diverse data sources in a unified way. Developers can reuse, configure, collaborate, and orchestrate components from retrieved information to realize IoT applications without much effort, thereby increasing the interoperability and reusability of the TinyML ecosystem.

\begin{figure}[t]
  \centering
  \includegraphics[width=0.85\linewidth]{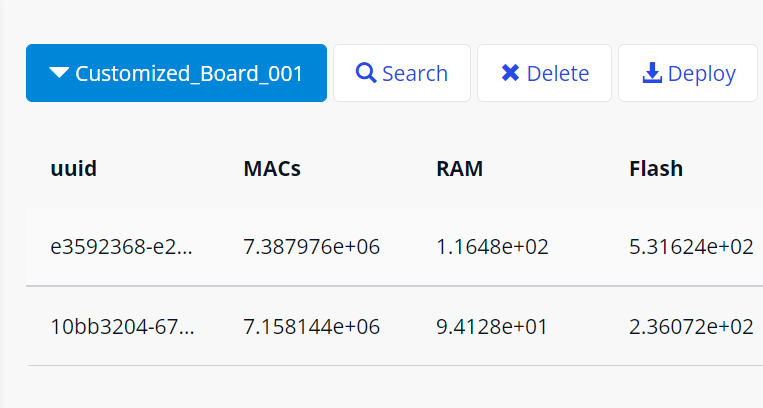}
  \caption{Integration of the concept in a low-code platform (query 2).}
  \label{mendix}
\end{figure}

Furthermore, the learning curve for Semantic Web technologies is not flat. Therefore, we proposed framework integration as a back-end service into a low-code platform and hiding the underlying mechanisms through a user-friendly GUI. Thus, users can fill out the provided templates with their requirements and get results with only a few clicks. For presentation, we implement the concept in a low-code platform, shown in Figure \ref{mendix}. Information is parsed in the background after selecting the IoT device described in query 2, and a corresponding SPARQL query is immediately formulated to fetch a set of possible ML models. The user can select one desired model and click the "Deploy" button, which initiates the (mass) deployment of the selected ML model to desired devices without interrupting the production process. Interested readers can refer to our repository, where we provide an example of deploying an ML model to an IoT board on the fly using Bluetooth.

\section{Conclusion and Future Work}
\label{section:conclusion}

TinyML brings intelligence to embedded devices and opens the door toward advances, yet co-managing disparate components (ML models and IoT devices) remain a challenge. In this work, we supplement TinyML with Semantic Web technology, making the knowledge of a TinyML system scalable, sharable, and interpretable, which is essential for keeping pace with the booming development of IoT. We presented an ontology schema tailored for neural network models in TinyML. The proposed ontology aligned with the W3C WoT TD ontology, which semantically describes IoT devices. These semantic schemas serve as information models for extracting and formulating heterogeneous information of a TinyML system in a unified way. We contributed tools and methods for constructing, managing, and using a TinyML KG. Furthermore, the effectiveness and correctness of the method were evaluated on 23 publicly available ML models and six embedded devices, where the extracted knowledge is stored in a graph database that humans and machines can consume. Three case studies were introduced to validate the concept addressing different industrial management problems in TinyML. Finally, we recapped the workflow of the framework and illustrated how this concept could be brought into a low-code platform, allowing every IoT developer to quickly get into the game and manage their TinyML system at scale.

Our future work includes: (1) Provisioning the toolchain and making the framework ready for use by anyone (2) Adapting the ontology toward different use cases. For example, we can extend the semantic schema of sensors with specific hardware settings, such as sampling rate, to realize more fine-grained knowledge management. (3) Currently, we provide the code for semantic modeling of ML model trained using Tensorflow Lite Microcontroller. With other embedded libraries for TinyML available, there is a desire for the framework and TinyML-as-a-service~\cite{Doyu2021} synergy, which allows developers to compile and deploy ML models using query results regardless of the underlying runtime platform. (4) Recently, there has been an increasing interest in graph neural networks due to the success of ML and KG. Since a TinyML system is inherently a graph, knowledge-infused ML methods can enhance an IoT system in various ways, such as anomaly detection in a large-scale network. Therefore, we will strive for other novel ML methods to intelligently manage TinyML.

\bibliographystyle{ACM-Reference-Format}
\bibliography{sample-base}

\appendix

\end{document}